\documentclass[conference]{IEEEtran}
\IEEEoverridecommandlockouts
\usepackage{cite}
\usepackage{amsmath,amssymb,amsfonts}
\usepackage{algorithmic}
\usepackage{graphicx}
\usepackage{textcomp}
\usepackage{xcolor}
\def\BibTeX{{\rm B\kern-.05em{\sc i\kern-.025em b}\kern-.08em
    T\kern-.1667em\lower.7ex\hbox{E}\kern-.125emX}}
\usepackage[english]{babel}
\usepackage{comment}

\usepackage[bottom=25.4mm,top=19.1mm,left=15.9mm,right=15.9mm]{geometry}

\begin{document}

\title{Hybrid Human-Machine Perception via Adaptive LiDAR for Advanced Driver Assistance Systems}

\author{\IEEEauthorblockN{1\textsuperscript{st} Federico Scarì}
\IEEEauthorblockA{\textit{Cognitive Robotics} \\
\textit{Delft University of Technology}\\
Delft, The Netherlands \\
f.scari@tudelft.nl}
\and
\IEEEauthorblockN{2\textsuperscript{nd} Nitin Jonathan Myers}
\IEEEauthorblockA{\textit{Systems and Control} \\
\textit{Delft University of Technology}\\
Delft, The Netherlands \\
n.j.myers@tudelft.nl}
\and
\IEEEauthorblockN{3\textsuperscript{rd} Chen Quan}
\IEEEauthorblockA{\textit{Systems and Control} \\
\textit{Delft University of Technology}\\
Delft, The Netherlands \\
c.quan@tudelft.nl}
\and
\IEEEauthorblockN{4\textsuperscript{th} Arkady Zgonnikov}
\IEEEauthorblockA{\textit{Cognitive Robotics} \\
\textit{Delft University of Technology}\\
Delft, The Netherlands \\
a.zgonnikov@tudelft.nl}
}
\maketitle

\begin{abstract}
Accurate environmental perception is critical for advanced driver assistance systems (ADAS). Light detection and ranging (LiDAR) systems play a crucial role in ADAS; they can reliably detect obstacles and help ensure traffic safety. Existing research on LiDAR sensing has demonstrated that adapting the LiDAR's resolution and range based on environmental characteristics can improve machine perception. However, current adaptive LiDAR approaches for ADAS have not explored the possibility of combining the perception abilities of the vehicle and the human driver, which can potentially further enhance the detection performance. In this paper, we propose a novel system that adapts LiDAR characteristics to human driver's visual perception to enhance LiDAR sensing outside human's field of view. We develop a proof-of-concept prototype of the system in the virtual environment CARLA. Our system integrates real-time data on the driver’s gaze to identify regions in the environment that the driver is  monitoring. This allows the system to optimize LiDAR resources by dynamically increasing the LiDAR’s range and resolution in peripheral areas that the driver may not be attending to. Our simulations show that this gaze-aware LiDAR enhances detection performance compared to a baseline standalone LiDAR, particularly in challenging environmental conditions like fog. Our hybrid human-machine sensing approach potentially offers improved safety and situational awareness in real-time driving scenarios for ADAS applications.
\end{abstract}

\section{Introduction}
\IEEEPARstart{T}{he} development of intelligent vehicles and advanced driver assistance systems (ADAS) has revolutionized the automotive industry and will shape the future of mobility \cite{b1, b2}. ADAS promise to significantly enhance driver safety, convenience, and overall vehicle performance. A key component of these systems is LiDAR (Light Detection and Ranging), which provides 3D point cloud data representing obstacles, road users such as pedestrians, vehicles, and other objects in the vehicle's surroundings \cite{LiDAR_overview_percept}. Traditional LiDAR systems operate with fixed sensing parameters that do not account for the varying conditions encountered in real-world driving scenarios. For instance, in the case of spinning LiDARs, the instantaneous rotational speed is typically fixed over a rotation. This results in a uniform angular resolution over the LiDAR's 360-degree field of view, which may be suboptimal \cite{b3}. In practice, dynamically adapting the instantaneous rotational speed over the field of view enables the LiDAR to vary its angular resolution over different regions to optimize its detection performance. 

\subsection{Related work}
Prior work has developed methods for adapting LiDAR sensing to external environmental factors. In \cite{b3, b4, b5, b6}, the point cloud density is increased in regions of interest (RoI), with the RoI typically determined through external camera systems. For example, in \cite{b3}, moving targets are identified within the RoI by co-locating the LiDAR with a camera that shares the same field of view. In \cite{b4}, the RoI is based on probable pedestrian coordinates identified from sparse LiDAR scans. In \cite{b5}, regions of high activity are detected using an event camera to define the RoI. Additionally, \cite{b6} optimized both sampling coordinates and the inpainting algorithm for images captured by a co-located camera. More recently, \cite{b7} proposed a system that adapts both the range and resolution of the LiDAR over the field of view. However, the adaptations in \cite{b7} are performed using offline location-based static topology maps, without considering real-time environmental conditions.

In parallel with the progress in adaptive LiDAR systems, intelligent driver monitoring systems (DMS) have been developed for ADAS to ensure the attentiveness of the human driver (who is fully or partially in responsible for the driving task). DMS often center around detecting the gaze direction of the driver. For instance, prior work has leveraged driver’s gaze within distraction warning systems \cite{b8, b9, b10, b11}. These systems monitor the driver’s gaze in real-time to issue alerts when the driver is deemed distracted. 

Overall, both adaptive LiDARs and driver monitoring systems are being actively used in advanced driver support systems. However, there is a lack of research exploring the potential for combining LiDAR with DMS. In particular, existing adaptive LiDAR systems for ADAS do not take into account information on human driver's gaze, which leads to LiDAR perceiving often the same objects as the human. This duplication might provide for better reliability of the joint driver-vehicle perception but risks missing out on improvements of accuracy of LiDAR sensing in the areas that are not  perceived by the driver. At the same time, DMS can readily provide the LiDAR with the information on regions currently perceived by the driver, such that the LiDAR can complement the driver's sensing instead of duplicating it. 

\subsection{Contributions}
This paper investigates the potential of hybrid human-machine perception for ADAS by combining LiDAR and human driver gaze tracking. 
The main contributions of our work are i) the novel design of a system that adaptively changes the LiDAR characteristics based on real-time human driver gaze information, and ii) the proof-of-concept implementation of gaze-aware LiDAR in a simulated environment CARLA and its evaluation. The integration of gaze information allows our system to determine the driver's region of focus and adjust the LiDAR parameters in real time while maintaining the same average rotational speed and the same average laser power consumption as a traditional, non-adaptive LiDAR. Our proposed design optimizes the LiDAR’s sensing range to minimize inefficient overlap between the LiDAR's field of view and the areas the human driver is actively focusing on. By reducing the LiDAR's range or resolution within the driver’s focus, the system increases its attention to the areas the driver is not observing. This is achieved by enhancing the range and resolution of the LiDAR outside the driver’s FoV, potentially improving the human-ADAS system's ability to detect safety-critical objects in this region.

\section{Background}
\subsection{Power, Range, and Frame Rate in LiDAR Systems}
The range of a LiDAR is directly related to the power of the emitted laser pulse. Higher power results in a stronger signal, which can travel further and be reflected back from more distant objects. The maximum emitted power is determined by the admissible exposure limit \cite{AEL_ref}, which restricts the intensity of the emitted laser to prevent harm to human eyes. In practice, the intensity of the light received by the sensor decreases with the square of the distance. Therefore, to detect objects at longer distances, LiDAR systems must emit more power to compensate for this loss in signal strength.

The frame rate of a LiDAR system refers to how quickly the sensor can scan the scene, typically expressed in frames per second. For example, a typical automotive LiDAR has a frame rate of $20\, \mathrm{Hz}$ \cite{frame_Rate_ref}. Driving automation requires LiDAR systems to balance these two factors: achieving a high enough frame rate to achieve the desired perception while staying within power constraints to ensure safe and efficient operation. These constraints are particularly critical in battery-powered systems, such as electric vehicles, where energy efficiency is paramount.

\subsection{Resolution Control in LiDAR Systems}
A high angular resolution of the LiDAR aids detecting small objects or identifying the fine structure of a scene. Moreover, higher resolution enhances identification performance, enabling the LiDAR to more accurately distinguish objects on the road, thereby improving the system's ability to detect and classify various obstacles, such as vehicles, pedestrians, and road signs, with greater precision. In spinning LiDARs, the angular resolution is controlled by adjusting the rotational speed of the LiDAR's motor. A higher motor speed results in lesser time to scan an object, thereby leading to a poor angular resolution. In MEMS-based LiDARs, resolution control is achieved by varying the voltages applied to the micro-electromechanical system (MEMS) mirrors \cite{mems_Review}. 
\begin{figure*}[htbp]
    \centering
    \includegraphics[width=1\textwidth]{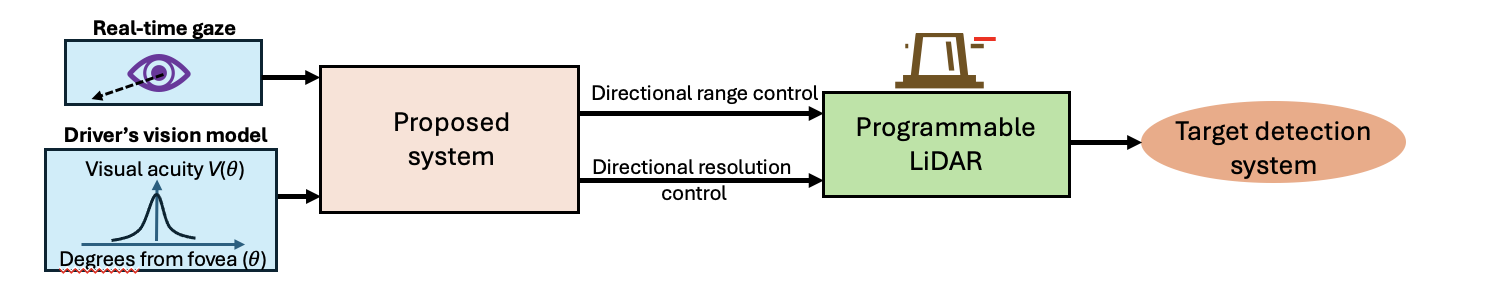}
    \includegraphics[width=1\textwidth]{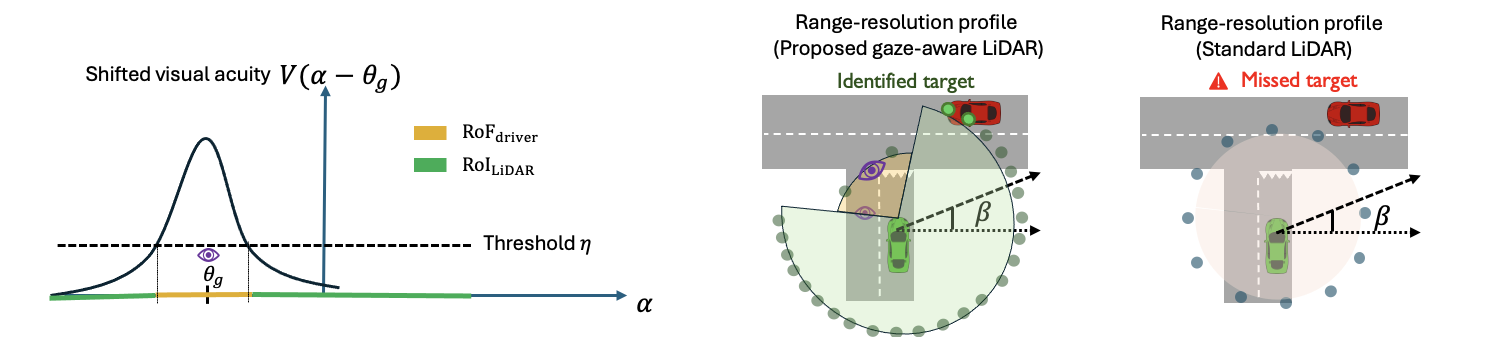}
    \caption{Overview of our proposed system. An implementation of the proposed system that determines the driver's region of focus $RoF_{\text{driver}}$, the region of interest for the LiDAR, from a shifted visual acuity function. The amount of shift is equal to the gaze direction ($\theta_{g}$). The proposed system achieves a higher range and a higher resolution outside $RoF_{\text{driver}}$ than a comparable standard LiDAR.}
\label{fig:system}
\end{figure*}

\section{System design and implementation}
To take advantage of the driver perception in ADAS, we propose a system with an adaptive spinning LiDAR that adjusts both its range and resolution based on real-time gaze information from the driver. This information is acquired by determining the driver’s region of focus ($RoF_{\text{driver}}$) using real-time gaze tracking. Using this information, the system dynamically modifies the LiDAR parameters to enhance sensing in regions outside the driver’s focus.

We consider a 2D LiDAR in our simulations. In our system, $RoF_{\text{driver}}$ is determined from the real-time gaze and the visual acuity function $V$ (Figure~\ref{fig:system}). The visual acuity function refers to the sharpness or clarity of vision, typically describing the eye's ability to distinguish fine details in its periphery~\cite{tunnacliffe1993}. This function can help define the driver's region of focus, representing the area where the driver can clearly detect objects. Assuming that the gaze direction $\theta_{g}$ (around which the region of focus (RoF) of the driver is centered) is known in real time, we can then define 
\begin{equation} 
RoF_{\text{driver}} = \left\{ \alpha : V(\alpha - \theta_{g}) > \eta \right\},
\end{equation}
where $\theta_{g}$ is the real-time gaze direction of the driver and $\eta$ is a threshold for visual acuity that is specific to an implementation. 
$RoF_{\text{driver}}$ then contains a range of directions that the driver is likely attending to. 

The region of interest for the LiDAR ($RoI_{\text{LiDAR}}$) is then defined as the complementary region of $RoF_{\text{driver}}$, i.e.,
\begin{equation}
RoI_{\text{LiDAR}} = [0, 2\pi) \setminus RoF_{\text{driver}}.
\end{equation}
Our goal is to enhance the range and resolution of the LiDAR in $RoI_{\text{LiDAR}}$ under an average laser power constraint and a frame rate constraint. Due to these constraints, our system puts less emphasis in scanning $RoF_{\text{driver}}$ as the driver already perceives this region well. The region outside the driver’s FoV, i.e., $RoI_{\text{LiDAR}}$, is emphasized in our LiDAR scan by increasing the range and resolution in $RoI_{\text{LiDAR}}$. With such an adaptation, our LiDAR improves detection in regions where the driver may not be actively monitoring. 

Our system uses two different ways to dynamically adapt the LiDAR in its $RoI$: a) range control, and b) resolution control; we also investigated simultaneous control of both range and resolution.

\subsection{Range Control}\label{RangeC}
Our LiDAR's laser employs two power levels, denoted by $P_{\text{high}}$ and $P_{\text{low}}$. While a low-power of $P_{\text{low}}$ is used in $RoF_{\text{driver}}$, a higher power of $P_{\text{high}}$ is used in the complementary region $RoI_{\text{LiDAR}}$ so that the average power over the field of view is same as a comparable standard LiDAR. For a scanning direction $\beta$ (Figure~\ref{fig:system}), our LiDAR's laser operates at a power of
\begin{equation}
P_{\text{LiDAR}}(\beta) = 
\begin{cases} 
P_{\text{low}}, & \text{if } \beta \in RoF_{\text{driver}} \\
P_{\text{high}}, & \text{if } \beta \in RoI_{\text{LiDAR}}
\end{cases}.
\end{equation}
At any given time, the power emitted by the LiDAR is adjusted depending on whether its current scanning direction is within the $RoF_{\text{driver}}$ or outside it ($RoI_{\text{LiDAR}}$). For an average power of $P$ with a standard non-adaptive LiDAR, the power levels in our system are chosen such that

\begin{equation}\label{equation powewrrange}
\frac{\Delta_{\text{Driver}}P_{\text{low}}+\Delta_{\text{LiDAR}}P_{\text{high}}}{2\pi}=P,
\end{equation}
where $\Delta_\text{\text{Driver}}$ is the angular width of the driver’s region of focus, equivalently the portion of the LiDAR’s field of view where the driver is actively looking. Our LiDAR emits a lower power $P_{\text{low}}$ in this region because the driver is already monitoring it. $\Delta_{\text{LiDAR}}$ is the angular width of the region outside the driver’s focus ($RoI_{\text{LiDAR}}$), equivalently the region that demands an increased attention from the LiDAR. Our LiDAR emits a higher power $P_{\text{high}}$ to enhance the LiDAR's detection capability in these peripheral areas. Finally, $2\pi$ represents the total angular span of the LiDAR's field of view (360 degrees).

\subsection{Resolution Control}\label{ResolutionC}
We use $\omega$ to denote the rotational speed of a standard spinning LiDAR, equivalently the corresponding frame rate is $\omega/2\pi$. A standard spinning LiDAR uses a constant instantaneous rotational speed to result in the same angular resolution over the field of view. Such a rotational speed profile does not account for gaze awareness. 

The spin rate of the proposed LiDAR is increased (denoted by $\omega_\text{high}$) within the driver’s focus to minimize unnecessary scanning. This results in a lower angular resolution within $RoF_{\text{driver}}$. The spin rate is reduced (denoted by $\omega_\text{low}$) outside the driver’s focus to improve resolution:
\begin{equation}
\omega_{\text{LiDAR}}(\beta) = 
\begin{cases} 
\omega_\text{high}, & \text{if } \beta \in RoF_{\text{driver}} \\
\omega_\text{low}, & \text{if } \beta \in RoI_{\text{LiDAR}}
\end{cases}.
\end{equation}
Due to such spin rate adaptation, the proposed LiDAR enhances resolution outside the driver’s focus, ensuring that areas outside the driver’s field of view are scanned more thoroughly. For a fair comparison with a standard LiDAR operating at a uniform spin rate of $\omega$, the time needed to scan $RoF_{\text{driver}}$ and the time needed to scan $RoI_{\text{LiDAR}}$ must sum to $2\pi/ \omega$, i.e.,
\begin{equation}
\frac{\Delta_{\text{Driver}}}{\omega_\text{high}}+
\frac{\Delta_{\text{LiDAR}}}{\omega_{\text{low}}}=\frac{2\pi}{\omega}.
\end{equation}
\subsection{Proof-of-concept implementation}
To demonstrate the feasibility and effectiveness of the proposed LiDAR system, we implemented a proof-of-concept in a simulation environment. The system was tested using the CARLA simulator \cite{b12}, an open-source urban driving simulator widely used for autonomous driving research. CARLA allows us to simulate a variety of driving conditions, including clear and foggy weather, which helps evaluate the system’s performance in real-world-like scenarios. It also provides the ability to track the vehicle’s movements and environmental variables in real-time, with LiDAR point clouds generated dynamically based on the simulated driving conditions.

\begin{figure*}[htbp]
    \centering
    \includegraphics[width=1\textwidth]{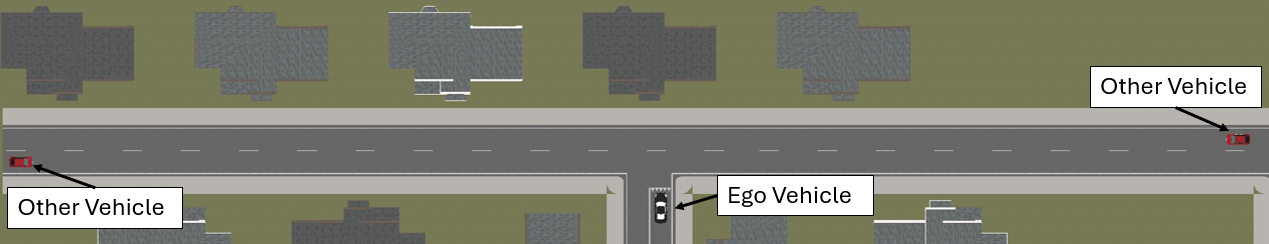}
    \caption{The simulation scenario used in our evaluation. The ego vehicle (black vehicle) stands at a T-cross intersection and two other vehicles approach from the left and from the right.}
    \label{fig:track}
\end{figure*}

For the baseline LiDAR, we used the LiDAR sensor model available in the CARLA simulator. The CARLA simulator provides a realistic virtual environment with a LiDAR sensor that can simulate various driving scenarios, including both static and dynamic obstacles. The LiDAR in CARLA operates with a 360-degree field of view and a fixed range of 100 meters and 64 channels to capture detailed point clouds. The system operates at a frequency of 20 Hz, generating approximately 500,000 points per second. This LiDAR model serves as the control system in our experiments, allowing us to compare its performance with our adaptive gaze-aware LiDAR system.

Real-time gaze data was measured using the Varjo VR3 headset, which is equipped with advanced eye-tracking technology. The VR3 headset provides high-precision gaze tracking, allowing for accurate real-time monitoring of the driver’s focus. The gaze data is continuously fed into the system, enabling dynamic adjustments to LiDAR parameters based on the driver’s attention. This integration ensures that the LiDAR’s performance is adapted in real-time.

In our system, the visual acuity function is used to model the driver's field of view, representing the area where the driver can detect objects with the highest clarity. For simplicity, we define the $RoF_{\text{driver}}$ as a 60-degree field of view. This means the driver is assumed to focus sharply within a 60-degree angle, while peripheral areas outside this focus are less likely to receive the driver’s attention.

This proof-of-concept implementation provides a detailed and reproducible approach to integrating real-time gaze information with LiDAR sensing. The use of the CARLA simulator and the Varjo VR3 headset ensures that the system can be evaluated under realistic driving scenarios, and the methodology can be easily replicated for further research and development.

\section{System Evaluation}
\subsection{Experimental Setup}
We evaluated three variants of our gaze-aware LiDAR implementation (range control, resolution control, range and resolution control) against a baseline LiDAR in simulated environment under three fog conditions (0\%, 25\%, and 50\%).
We consider a T-cross intersection scenario (Figure~\ref{fig:track}) with three vehicles: the ego vehicle is entering the intersection in the presence of two other vehicles: one approaching from the left and another from the right. 

In this proof-of-concept evaluation, our aim was to investigate the performance of the system in a hypothetical scenario where the ego vehicle driver does look to the left but does not look to the right; hence the vehicle needs to detect the vehicle on the right through the LiDAR, in order for ADAS to warn the driver as early as possible. For reproducibility, we did not involve actual human drivers in the evaluation at this stage but instead used a fixed driver gaze direction (looking to the left at all times) in order to reliably measure the system performance in a given scenario.
The detection of the vehicle on the right was implemented based on the point cloud generated by the LiDAR with a heuristic algorithm (we assumed the vehicle was detected as soon as the LiDAR detected the first point at the expected vehicle height). 

We evaluated the system's performance using two metrics: time-to-arrival (TTA) of the right vehicle at which it was detected and LiDAR point cloud density.
Here, TTA was defined as the time in which the right vehicle is expected to arrive to the intersection (moving with the constant speed of 50km/h), measured at the moment it was detected by the LiDAR. A higher TTA-at-detection is desired as it provides the driver with more time to react to the oncoming vehicle outside the driver’s FoV. To further illustrate the detection performance, we analyzed the point cloud density of the LiDAR outside the driver's FoV. Higher densities enable the LiDAR to detect targets not only earlier but also more reliably.

\subsection{Evaluation of Detection Timing}

\begin{figure*}[htbp]
    \centering
    \includegraphics[width=1\textwidth]{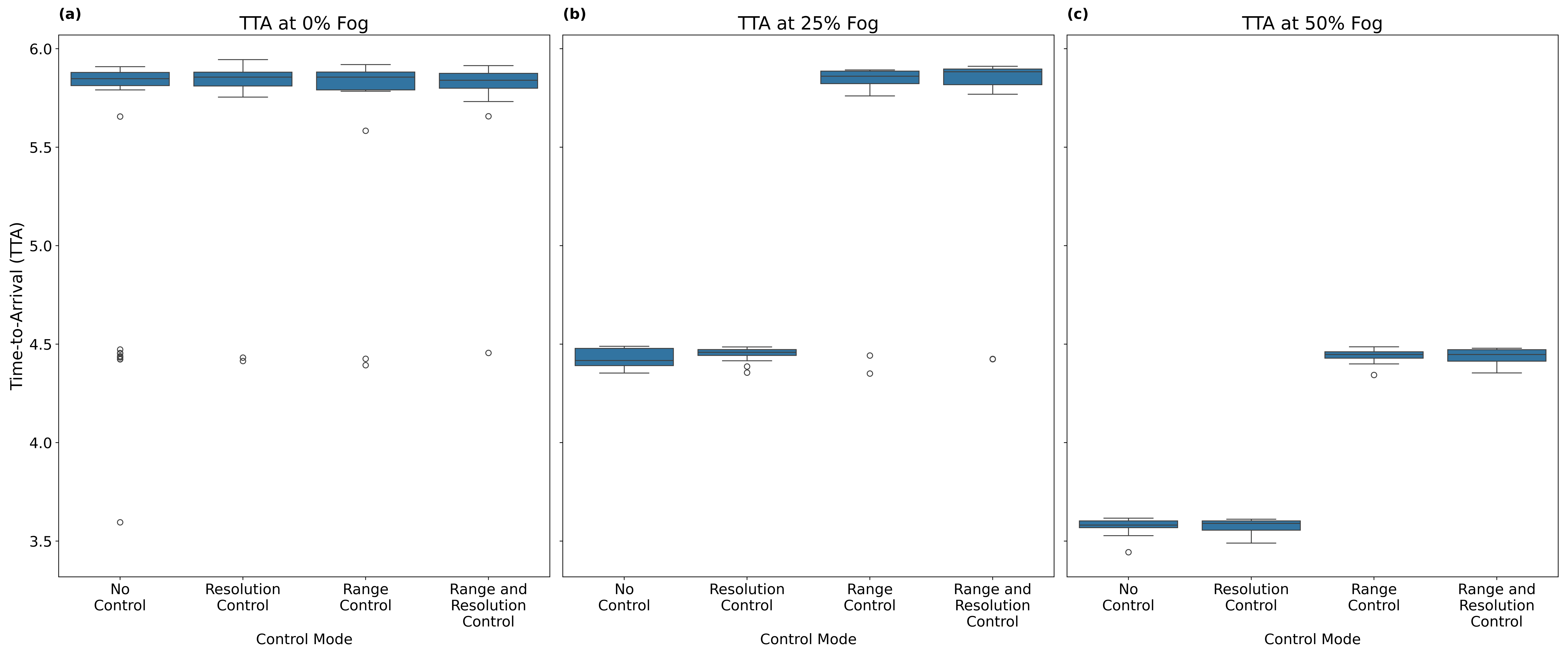}
    \caption{Time-to-arrival (TTA) comparison (in seconds) for four evaluated LiDAR control methods across three fog levels.}
    \label{fig:TTA}
\end{figure*}

\begin{figure*}[h!]
    \centering
    \includegraphics[width=1\textwidth]{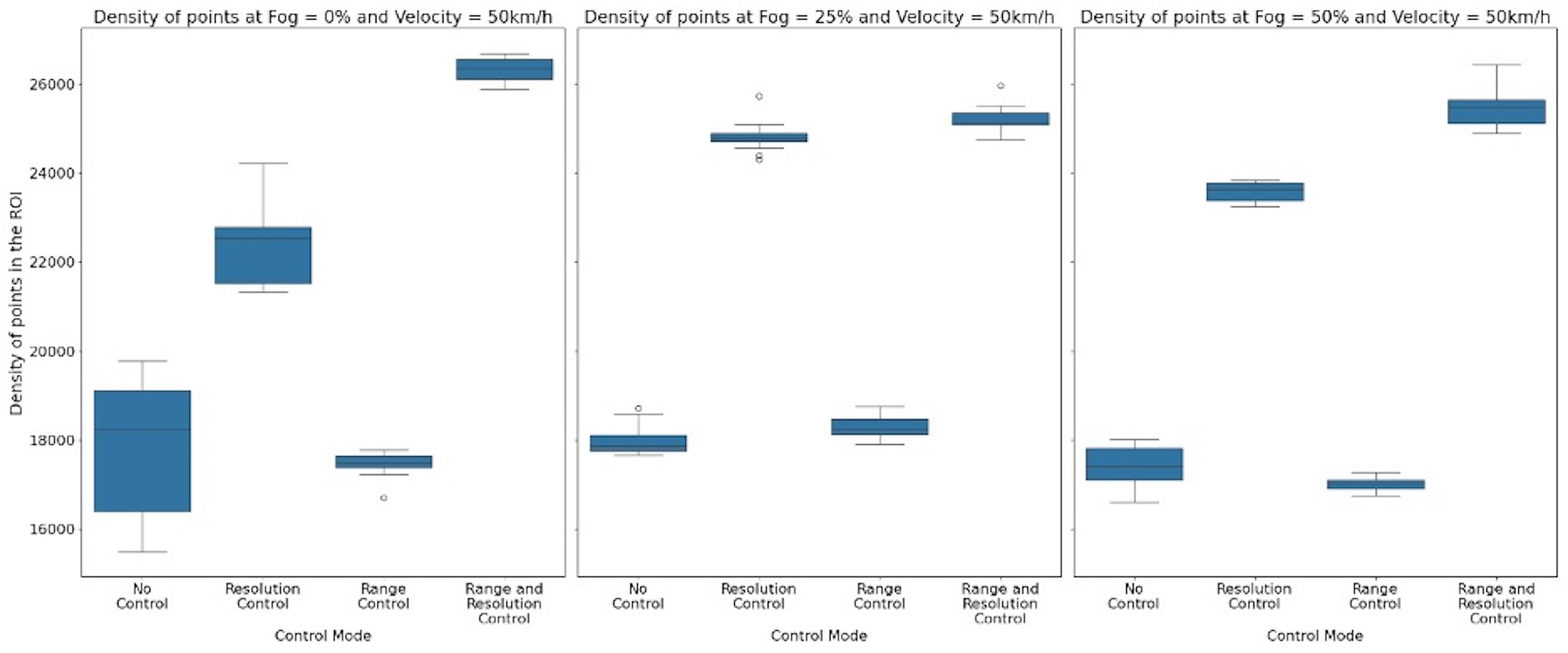}
    \caption{Density of LiDAR data points in the LiDAR's region of interest for four evaluated LiDAR control methods across three fog levels.}
    \label{fig:density}
\end{figure*}

Figure~\ref{fig:TTA} illustrates the differences in measured TTA across the three evaluated adaptive LiDAR variants, as compared to the standard, non-adaptive LiDAR. In the no-fog condition, the right vehicle was detected at TTA values ranging from approximately 4 to 6 seconds, with the median TTA around 5.8s. Here, all four LiDARs show very similar performance in terms of detection timing. 
With the fog density of 25\%, TTA values decrease for the standard and the resolution-controlled LiDARs, ranging around 4.5 seconds (Figure~\ref{fig:TTA}b). Notably, range-controlled and the range-and-resolution-controlled gaze-aware LiDARs demonstrated improved efficiency, achieving a TTA value comparable to that in the no-fog condition. 
In the scenario with 50\% fog (Figure~\ref{fig:TTA}c), TTA values further declined, showing even later detection of the vehicle, with TTA ranging from 3.5 to 4.5 seconds. The two LiDARs including the range control consistently yielded the best performance, detecting the right vehicle approximately 1 second earlier compared to the non-adaptive and the resolution-controlled gaze-aware LiDAR. 
 
Overall, the TTA evaluation results illustrate that in simulated foggy weather conditions, our gaze-aware LiDAR with range control can detect the approaching target outside of driver's FoV substantially earlier compared to the baseline, non-adaptive LiDAR. At the same time, controlling the LiDAR resolution doesn't yield any benefits in detection timing.

\subsection{Evaluation of Point Cloud Density}
Figure~\ref{fig:density} illustrates the results of our analyses of the density of data points collected in the LiDAR's region of interest. In the absence of fog (Figure~\ref{fig:density}a), the density of data points ranged from approximately 16,000 to over 26,000 points per degree. Unlike the analyses of detection timing, here we found that under 0\% fog the LiDARs differed in their point cloud densities. In particular, the gaze-aware LiDAR with range and resolution control achieved the highest median densities, followed by the resolution-controlled LiDAR. The range-controlled and the baseline non-adaptive LiDAR had the lowest point cloud densities. As we increased the fog density to 25\% (Figure~\ref{fig:density}b) and 50\% (Figure~\ref{fig:density}c), the density of data points did not change substantially. Similar to the no-fog condition, the two gaze-aware LiDARs with resolution control achieved high point cloud densities. 

Overall, our analyses revealed substantial benefits of adapting the LiDAR resolution and range depending on the driver's gaze direction. The effects included increased point cloud density (across all studied weather conditions) with resolution control and faster target detection in foggy conditions with range control. Our findings highlight that incorporating gaze information can enhance the perception of the hybrid human-vehicle system, especially in adverse weather conditions.

\section{Limitations and future work}
While our system leverages real-time driver gaze information to enhance LiDAR sensing, it operates on the assumption that the areas the driver is focusing on are also perceived with sufficient attention. However, this assumption may not always hold true, e.g., during mind wandering~\cite{he2011}. Furthermore, in certain situations, such as low visibility conditions (e.g., fog), the driver may not perceive the environment with the same accuracy, making the system potentially less reliable in these circumstances. In such cases, the system should still continue improve the scanning outside the driver’s RoF. However, to account for potential inaccuracies in the driver’s perception, the system could reduce the difference in emitted power and resolution between the $RoF_{\text{Driver}}$ and $RoI_{\text{LiDAR}}$. This would allow for a more balanced reliance on both the driver’s gaze and the environmental data. This adjustment could be dynamically regulated, with thresholds for power and resolution being fine-tuned based on environmental factors, such as fog density, but also driver's situational awareness, ensuring that the system remains robust even when the driver’s attention might be compromised.

A limitation of the current prototype implementation is the simplicity of the heuristic algorithm used for vehicle detection. This detection method is simple and effective for the simulation environment, where the primary challenge is detecting a single vehicle approaching from the side. In this controlled environment, noise is already accounted for by the CARLA simulator, ensuring that the point cloud data is clean and suitable for detection tasks. As a result, the algorithm works effectively within this scenario, where only one vehicle is present. This is only a limitation of our proof-of-concept implementation but not the fundamental design of our system. Still, for real-world applications, appropriate detection methods need to be used to handle dynamic obstacles, multiple vehicles, and other sources of noise~\cite{mao2023}.

Lastly, another potential limitation is the risk of driver over-trusting the system. If the proposed LiDAR system performs well, drivers may start to over-rely on ADAS to warn them of potential threats, reducing their own vigilance and attention to the environment. This over-reliance could lead to a false sense of security, where the driver no longer engages in active scanning of their surroundings. To mitigate this, the system could incorporate periodic checks or alerts, ensuring that the driver remains attentive even when ADAS with the gaze-aware LiDAR is reliable.

\section{Conclusion}
This paper introduced a novel approach to enhancing LiDAR performance by incorporating real-time driver gaze data to adapt LiDAR parameters. Our system stands apart from existing methods by integrating the driver’s gaze directly with the LiDAR’s sensing capabilities. This real-time interaction ensures that the vehicle’s perception is continuously optimized. As a result, the system improves detection accuracy and overall safety, making the vehicle more responsive to potential hazards. In particular, the experimental evaluation in the CARLA environment showed that our system improved the performance of object detection, particularly in challenging conditions such as fog, and offered better situational awareness for the vehicle. These results are a first proof-of-concept demonstration of gaze-aware LiDARs outperforming a traditional LiDAR system, providing a promising advancement in vehicle sensing technology for automated driving.

\end{document}